**2021****Федеральное государственное автономное образовательное учреждение высшего образования
«Национальный исследовательский университет
«Высшая школа экономики»**

**Факультет компьютерных наук
Основная образовательная программа
Прикладная математика и информатика**

# КУРСОВАЯ РАБОТА
## Исследовательский проект на тему
## Оптимизации применения решающих деревьев с помощью SIMD инструкций

**Выполнил студент группы 186, 3 курса,
Миронов Алексей Дмитриевич**

**Руководитель КР:
кандидат физико-математических наук, руководитель группы инфраструктуры качества поиска
Хузиев Ильнур Масхудович****Москва 2021**

# Оглавление

## Abstract

Decision forest (decision tree ensemble) is one of the most popular machine learning algorithms. To use large models on big data, like document scoring with learning-to-rank models, we need to evaluate these models efficiently. In this paper, we explore MatrixNet, the ancestor of the popular CatBoost library. Both libraries use the SSE instruction set for scoring on CPU. This paper investigates the opportunities given by the AVX instruction set to evaluate models more efficiently. We achieved 35% speedup on the binarization stage (nodes conditions comparison), and 20% speedup on the trees apply stage on the ranking model.

## Аннотация

Решающий лес (ансамбль решающих деревьев) – один из наиболее популярных алгоритмов машинного обучения. Чтобы иметь возможность использовать сложные модели на большом объёме данных, например для ранжирования документов, нужно уметь быстро применять эти модели. В этой работе мы исследуем MatrixNet – предок популярной библиотеки CatBoost. В обоих библиотеках для применения на ЦП используется набор инструкций SSE. В рамках этой работы проверяется возможность ускорить применение за счёт AVX инструкций. Мы смогли ускорить модель ранжирования на 35% на стадии бинаризации (сравнения условий в узлах), и на 20% на стадии применения деревьев.

*Ключевые слова: решающий лес, SIMD, SSE, AVX, MatrixNet, CatBoost.*

## 1. Введение

Решающий лес – широко известный алгоритм машинного обучения. Среди наиболее популярных библиотек, реализующих модели на основе решающего леса, XGBoost и CatBoost, разрабатываемый в основном компанией Яндекс. Обе эти библиотеки используют метод градиентного бустинга для обучения моделей, они имеют по несколько тысяч отметок "Star" на GitHub [1] [2], их используют в разных сферах, с их помощью часто побеждают на соревнованиях по машинному обучению на Kaggle.

Скорость применения для моделей может быть критична, так как при большом объёме данных, стоимость вычислительных мощностей может быть слишком большой, а при ограниченных вычислительных мощностях или времени, возможно, что вместо модели, дающей наилучшее качество, придётся использовать модель, укладывающуюся в ограничения по ресурсам. Одним из классических примеров, где подобная проблема встречается на практике является проблема ранжирования, где нужно для каждого поискового запроса вычислять релевантности для набора кандидатов, число которых может быть велико, а их уменьшение может сказываться на качестве результатов, и при этом имея бюджет в доли миллисекунд времени центрального процессора для каждой пары запрос-документ. В таких условиях даже небольшое изменение производительности может сильно сказаться на результате.

Одним из ключевых особенностей CatBoost является высокая производительность, которую удаётся достичь за счёт использования решающих деревьев специального вида (oblivious-tree, смотрите подробности ниже). Альтернативной реализацией применения решеающих деревьев подобной структуры является более старая разработка той же компании Яндекс – Matrixnet. Она за счёт незначительного на практике проигрыша в точности выиграват в скорости применения около 10%. В связи с этим, далее в работе мы будем использовать именно эту реализацию (MatrixNet) как базовую в наших сравнениях.

В рамках этой задачи мы считаем, что модель решающего леса, используемая для ранжирования документов, получает на вход вектор **вещественных признаков** $FACTOR$ для каждого документа и возвращает одно число для каждого документа (релевантность). В узлах деревьев записаны условия вида $FACTOR_{INDEX_i} < THRESHOLD_i$, где $INDEX, THRESHOLD$ – векторы модели, а $i$ – номер узла. Результат вычисления этих условий мы будем называть **бинарными признаками**. Чтобы получить ответ, на каждом шаге, начиная от корня дерева, мы, в зависимости от значения бинарного признака (однозначно определяется по номеру дереву и расстоянию от корня), спускаемся в одно из двух поддеревьев, пока не достигнем листа дерева, в котором записан ответ. Итоговой ответ считается как сумма ответов по всем деревьям.

В рамках этой статьи исследуется оптимизация скорости с помощью Intel SIMD (Single Instruction Multiple Data) – инструкций процессора, позволяющих одновременно выполнять одну операцию над несколькими числами. Текущая реализация MatrixNet написана на SSE2, выпущенном в 2001 году, и имеющем 16 128-битных регистров и 214 инструкций. Сейчас уже выпущен и доступен на многих процессорах AVX-512, имеющий 32 512-битный регистров и больше 5000 инструкций [7]. Нужно выяснить, получится ли с помощью новых регистров и инструкций добиться большей производительности. Также мы ставим своей целью исследовать влияние различных типов оптимизаций на скорость в данной задаче.

Вторая глава посвящена исследованию литературы и реализации MatrixNet, третья – погружению в SIMD и Intel Architectures Optimization Manual, четвёртая – методологии исследования и тестирования, пятая – идеям и экспериментам с бинаризацией (вычислением значений бинарных признаков), шестая – идеям и экспериментам с алгоритмами применения, в седьмой подведены итоги работы.

# 2. Существующие работы и решения

При применении стандартного дерева есть проблемы, вызванные ветвлением. При наивном обходе дерева мы не можем знать, через какие узлы мы пройдём, что не даёт нам заранее загружать данные в кэш для следующих вершин и распараллеливать процесс. Разные подходы отличаются тем, как обходить решающие дерево.

## 2.1. V-QuickScorer

Статья итальянского университета "ISTI-CNR and Istella Srl" 2016 года [4] рассматривает использование SSE и AVX-2 для ускорения алгоритма QuickScorer [3]. Сначала из всех вершин дерева выписываются условия вида $FACTOR_{INDEX_i} < THRESHOLD_i$, и для каждого условия записывается битовая маска листьев, которые будут недоступны, если условие ложно (то есть нули в маске будут только на отрезке поддерева, в которое мы не пойдём, если дойдём до этого условия, и оно окажется ложно). Набор ложных условий однозначно задаёт путь к листу с ответом, поэтому достаточно определить все ложные условия и сделать побитовую конъюнкцию всех их масок, тогда первый бит в маске (или последний, в зависимости от того, влево или вправо идём, если условие ложно) будет задавать ответ дерева. Чтобы эффективно посчитать конъюнкцию, эти условия группируются по $INDEX$ и сортируются по $THRESHOLD$, тогда для каждого документа для каждого числового признака нужно брать конъюнкцию только на некотором префиксе условий в силу сортировки.

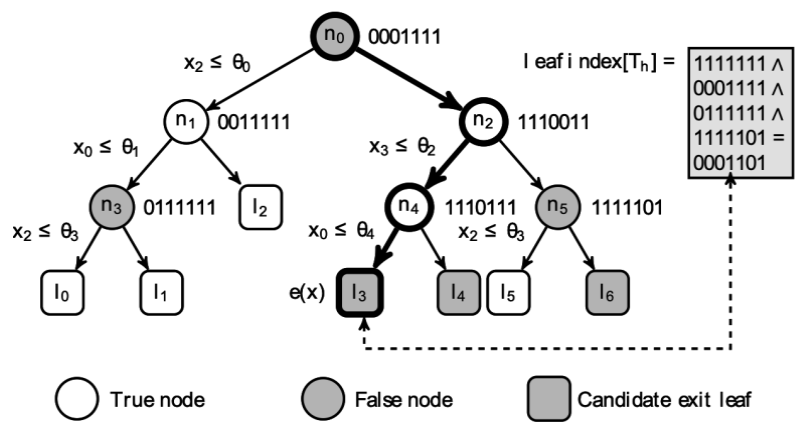

*Рисунок 2.1 демонстрация V-QuickScorer в статье Microsoft Research [5]*

Здесь используются SIMD инструкции для параллельного сравнения чисел и для работы с битовыми масками. В эксперименте процент ускорения сильно зависит от данных, но в целом использование SSE давало ускорение примерно в 2 раза, а AVX-2 примерно в 2.4 по сравнению с реализацией без SIMD инструкций.

## 2.2. RapidScorer

У QuickScorer есть существенный недостаток: время для одного документа линейно зависит от числа листьев дерева, то есть при дереве глубины $n$ сложность получается $O(2^n)$. Это не представляет проблем, пока используются неглубокие деревья, а в решающем лесу, построенным градиентным бустингом, часто используют неглубокие деревья. Но в Microsoft Research на наборе данных Bing Ads показали, что глубокие деревья могут существенно повысить качество ранжирования. В 2018 году в Microsoft Research выпустили статью про RapidScorer [5] – модификацию V-QuickScorer, которая улучшала время применения одного дерева до $O(\sqrt{2^n})$. Сделано это было с помощью специального кодирования битовой маски под названием эпитом, которая позволяла хранить меньше байт, но при этом всё ещё использовать возможности SIMD инструкций. Можно заметить, что в V-QuickScorer битовая маска в вершинах дерева представляет из себя отрезок из нулей (то есть сначала идут только единицы, потом только нули, потом снова только единицы). При этом конъюнкцию бит с единицами можно пропустить, в результате конъюнкции с битовой маской изменяются только биты, которым соответствует ноль в маске. Приведённая в статье теорема утверждает, что нулей в среднем $O(\sqrt{2^n})$. Можно проводить конъюнкцию только для этого отрезка из нулей. Этот отрезок может иметь разную длину, но достаточно выполнить конъюнкцию только на байтах на границах этого отрезка, в середине отрезка байты битовой маски нулевые и мы можем сразу записать ноль.

RapidScorer – вероятно текущий state-of-the-art по скорости применения. За счёт этой и некоторых других оптимизаций он показывает результаты примерно в 1.3 раза лучше V-QuickScorer даже при 32 листьях, а XGBoost он обгоняет до 30 раз при 8 листьях, но всего в 10 раз при 256 листьях.

### 2.3. Matrixnet

Matrixnet и CatBoost – библиотеки машинного обучения, разрабатываемые Яндексом. Matrixnet был проектом с открытым исходным кодом, но сейчас в публичном поле его заменил CatBoost, который во многом похож на MatrixNet. Обе библиотеки работают с oblivious-tree.

Деревья в Matrixnet высоты $n$ задаются $n$ условиями и $2^n$ ответами. То есть в отличии от стандартных решающих деревьев, здесь на одинаковом расстоянии от корня в узлах дерева стоят одинаковые условия. Такая структура дерева делает несовместимым модели MatrixNet с моделями большинства других библиотек и не позволяет сравнивать его с ними по скорости. Однако это позволяет применять деревья намного быстрее, что позволяет иметь больше деревьев в модели.

Такая структура деревьев даёт возможность получить ответ дерева для документа, взяв используемые в дереве признаки, посчитанные для документа, записав их в виде двоичного числа и взяв ответ из массива ответов дерева, используя получившиеся число в качестве индекса. Например, если для дерева высоты 3 вычисленные признаки: true, false, true, то ответ будет лежать в массиве по индексу 5.

Для простоты можно разделить этот процесс на два этапа:

- **Бинаризация**, во время которой вычисляются сами бинарные признаки для каждого документа, то есть выполняются сравнения вида $FACTOR_{INDEX_i} < THRESHOLD_i$. Легко посчитать, что максимальное количество бинарных признаков может быть равно сумме высот деревьев, но фактически в моделях их меньше, так как при обучении модели сначала строятся бинарные признаки, а потом по ним строятся деревья, и один признак может использоваться в большом числе деревьев. Число бинарных признаков является одним из важных параметров модели.
- **Применение деревьев**, во время которого по бинарным признакам считаются индексы для массива ответов каждого дерева, и складываются числа из массивов ответов, для получения финального ответа.

## 3. Intel Cascade Lake SIMD

SSE и AVX добавляют регистры большего размера, по сравнению с x86-64, где регистры общего назначения являются 64-битными. В зависимости от расширений процессора, могут быть доступны регистры разного размера.

*Таблица 3.1 Размеры и количества регистров в зависимости от расширения процессора [7]*

| Расширение процессора | Размер регистра | Количество соответствующих регистров |
|---|---|---|
| SSE-SSE4.2 | 128 | 16 |
| AVX, AVX2 | 128 | 16 |
| AVX, AVX2 | 256 | 16 |
| AVX-512VL | 128 | 32 |
| AVX-512VL | 256 | 32 |
| AVX-512F | 512 | 32 |

Так же начиная с AVX-512 появился новый вид AVX регистров – opmask [11]. Были добавлены восемь 64-битных opmask регистров, которые используются некоторыми AVX-512 инструкциями, чтобы принимать или возвращать 8, 16, 32 или 64 битов информации.

Для удобства разработки, Intel предоставляет библиотеку интринсиков, позволяющую работать с SIMD регистрами и инструкциями из кода на C/C++ [7].

Инструкции процессора пишутся последовательно и в предположении, что они будут выполнятся последовательно, но фактически несколько инструкций может выполняться одновременно [6]. Ядро процессора в Cascade Lake, имеет несколько портов, у каждого порта есть свой набор инструкций, которые он может выполнять. Если две инструкции можно расположить на разных портах, и результат второй инструкции не зависит от результата первой, то они могут выполняться одновременно, поэтому важно писать код, в котором инструкции могут выполняться параллельно. Документация Intel для каждой линейки процессоров описывает производительность каждой инструкции с помощью двух параметров: latency – число тактов процессора, требующихся на выполнение одной инструкции, и throughput – как часто в тактах инструкция может стартовать (latency, делённый на число инструкций, которые могут выполняться одновременно) [8].

Чтобы операции могли распараллеливаться, нужно, чтобы подряд идущие инструкции использовали разные регистры, при этом важно, чтобы код не использовал больше регистров, чем доступно, так как это может привести к тому, что компилятор будет выгружать промежуточные результаты в память, чтобы освободить регистры [4].

К тому же важно, чтобы в коде было как можно меньше ветвлений, чтобы процессор знал, какая инструкция идёт следующей. Для этого используется разворачивание циклов и отказ от условий в "If", которые нельзя вычислить во время компиляции [6].

Как будет показано дальше, для нас так же критична производительность работы с памятью. Помимо приведённых выше, руководство Intel по оптимизации даёт следующие советы:

- Операции выгрузки/загрузки в память должны укладываться в имеющуюся пропускную способность. На Cascade Lake можно загружать/выгружать до 512 бит одной инструкцией и выполнять до 2 таких инструкций одновременно [6].
- Выравнивать данные. В случае с 512 битными чтениями, нужно выравнивать по 64 байта [8] [10] [11].
- Использовать prefetch инструкции. Prefetch позволяет заранее указать, какие данные нужно загрузить в кэш [9].
- Соблюдать локальность данных. Кэш имеет иерархическую структуру и близко лежащие данные быстрее загружаются в кэш [9].

## 4. Методология исследования

Алгоритм применения решающих деревьев был разделён на бинаризацию (вычисление бинарных признаков) и применение документов (вычисление значений деревьев с последующим суммированием). Для каждой идеи был написан код, её реализующий. Также была реализована наиболее многословная неоптимизированная версия, чтобы быть эталоном для тестирования верности результатов и иллюстрации используемого в рамках данной задачи представления леса в памяти.

Используемая в качестве baseline реализация из MatrixNet в целях простоты восприятия была упрощена путём удаления не нужных в рамках данной работы возможностей библиотеки. Также, было сделано указанное выше разделение на бинаризацию и применение документов, которое в исходной реализации отсутствовало

Для проверки написанного кода, был реализован тестовый стенд. На тестовом стенде установлен процессор Intel Xeon Gold 6230, который поддерживает все описанные выше расширения. Для измерения времени применения использовалась библиотека Google Benchmark. Google Benchmark измеряет процессорное время небольших по продолжительности операций. Процессорное время может отличаться от запуска к запуску из-за состояния кэша, поэтому для замеров проверялось, что средний квадрат отклонения в 100 раз меньше, чем среднее время (такая проверка не делалась в случае, когда замеры делаются для последовательного количества документов (1, 2, 3…1024), т. к. вычисления заняли бы несколько месяцев, а все ошибки и так видны на графики в виде внезапных всплесков).

Так же исследовалось, как на время влияет одновременная работа нескольких потоков применения. Каждый поток получал свою копию данных и модели.

Запускались алгоритмы как на практически используемой в Яндекс.Поиске модели, так и на нескольких синтетических моделях. Первая модель имеет 1950 вещественных факторов, 155377 бинарных признаков и 155100 деревьев, из которых 154995 высоты 6, остальные меньше. Синтетические модели позволяют показать, как по-разному ведут себя разные виды алгоритмов при разном соотношение параметров модели. В открытый доступ выложен только код для синтетических моделей.

Результаты как бинаризации, так и применения сравнивались друг с другом на полное равенство, чтобы убедиться, что все алгоритмы работают. Код написан так, чтобы собираться и корректно работать под всеми санитайзерами.

Код опубликован в репозитории: https://github.com/yandex/research-optimization-of-decision-tree-evaluation

# 5. Бинаризация

## 5.1. Наивная реализация

Формальная постановка задачи: на вход подаётся двумерный массив 32-битных вещественных чисел размера число документов на число вещественных факторов, нужно вернуть бинарные признаки в каком-нибудь формате.

Модель содержит массивы с описанием бинарных факторов и порогов. Массив с описанием содержит набор структур, каждая из которых задаёт индекс вещественного фактора ($INDEX_i$) и число бинарных факторов, ему соответствующих. Другой массив содержит пороги ($THRESHOLD_i$) – массив 32-битных вещественных чисел размера число бинарных факторов.

Для каждого документа считаем бинарные признаки, поддерживая текущий индекс бинарного фактора и индекс описания текущей группы бинарных факторов.

## 5.2. MatrixNet

В MatrixNet документы обрабатываются группами по 128 документов. Каждая группа дополнительно разделена на 8 подгрупп по 16 документов. Если число документов не кратно 128, то подгрупп может быть меньше. Если число документов не кратно 16, то считаться будет все равно так, как будто во всех подгруппах по 16 документов, но вместо нескольких последних документов, будет использоваться шум со стека.

Внешний цикл в MatrixNet идёт по массиву с описанием признаков, из этого массива берётся индекс текущего вещественного фактора. Дальше нужно загрузить вещественные факторы для всех документов группы, но признаки подаются в массиве, первым индексом которого является номер документа, а вторым – номер признака, то есть локально в памяти лежат разные признаки одного документа, а признаки с определённом индексом у группы документов разбросаны по памяти. Чтобы воспользоваться преимуществами кэша, участок этого массива размером 128 документов на 4 вещественных фактора транспонируется.

Далее по очереди для каждой подгруппы 32-битные вещественные факторы загружаются в SSE регистры, для одной подгруппы используется 4 128-битных регистра. Ещё один регистр используется для записи порогового значения для вычисления очередного бинарного признака. Производятся сравнения с помощью инструкции *cmpps*, она возвращает регистр, в котором для каждого числа в зависимости от результатов сравнения записаны либо 32 единицы, либо 32 нуля.

Записывать результаты в таком виде не оптимально, так как требует большого числа обращений к памяти, в то время, когда достаточно записать всего 128-бит, а обращения к памяти занимают дольше, чем большинство операций над данными, которые уже в регистрах. Для того, чтобы сжать это до 128 бит, сначала 4 регистра подгруппы сжимают до 1 с помощью последовательного применения инструкций *packssdw* и *packsswb* (упаковывают два регистра в один, беря из каждого регистра младшую половину из каждого блока по 32 или 16 бит). На выходе получается один регистр, в каждом из байт которого записаны либо 8 единиц, либо 8 нулей. Далее, чтобы избавиться от лишних единиц, выполняется конъюнкция с маской, в которой установлен последний бит каждого байта. Результат сдвигается влево на число бит, равное номеру подгруппы начиная от нуля. Итоговые 128-бит получаются дизъюнкцией результатов подгрупп.

Получается, что бинаризация для каждой группы возвращает массив длины число бинарных признаков на 16 байт. В каждом блоке из 16 байт лежат бинарные признаки для 128 документов, при этом лежат они не по порядку. Первому документу соответствует бит с нулевым смещением, второму – с восьмым, третьему с 16. А со смещением один находится бит, соответствующий семнадцатому документу. Чтобы сравнивать результат с другими алгоритмами бинаризации, упорядочить биты можно, используя следующую формулу: на смещении $i$ стоит бит, соответствующий документу номер $i \% 8 * 16 + i / 8 + 1$, где символ $*$ означает умножение, $/$ – деление с округлением вниз, а % – остаток от деления.

### 5.3. AVX реализация

Можно попробовать ничего принципиально не менять в алгоритме MatrixNet и развить описанные выше идеи, разбивая на группы размером 256 и 512, и подгруппы размером 32 и 64 документа, и используя наиболее близкие AVX инструкции для 256 и 512-битных регистров.

Но, инструкции упаковки, которые использовались, чтобы упаковать 4 регистра в один, работают по-другому, чтобы обеспечить совместимость с SSE версиями. Первые 128 бит четырёх регистров упаковываются в первые 128 бит результирующего регистра, вторые во вторые и так далее. Если в реализации для 128-битных регистров, соответствующий пятому документу бит находился на смещении 32, то в реализации для 256 или 512-битных регистров, он будет находится на смещении 128. Формулы для номера документа по смещению в массиве для больших размеров регистра принимают вид:

$i / 8 \% 4 + i / 8 / 4 * 8 \% 32 + i / 8 / 4 * 8 / 32 * 4 + i \% 8 * 32 + 1$, для 256

$i / 8 \% 4 + i / 8 / 4 * 16 \% 64 + i / 8 / 4 * 16 / 64 * 4 + i \% 8 * 64 + 1$, для 512

Так же для 512-битных регистров все инструкции сравнения возвращают результатом в opmask регистре, что добавляет ещё одну операцию на то, чтобы переместить результат сравнения в обычный.

### 5.4. Упорядоченный формат

В AVX-512 были добавлены инструкции сравнения для всех трёх размеров регистров, возвращающие результат в opmask регистрах. В таком виде нет необходимости выполнять сложные инструкции для упаковки результата, можно свести биты из нескольких регистров только сдвигами и дизъюнкциями. Более того, мы можем записывать биты для каждой подгруппы отдельно, поэтому нет смысла в больших группах, размером с размер регистра, и алгоритм может быть оптимальным для небольшого числа документов.

Ещё одним преимуществом является то, что на смещении $i$ стоит бит, соответствующий документу номер $i + 1$ и не нужно использовать сложные формулы для упорядочивания.

## 5.5. Транспонирование

С теоретической точки зрения было интересно, что будет, если убрать транспонирование из кода и передавать сразу транспонированный массив на вход. Внедрять подобное может иметь смысл, только если транспонирование уже где-то делается до этого в коде, и транспонирование можно убрать из двух мест.

## 5.6. Результаты

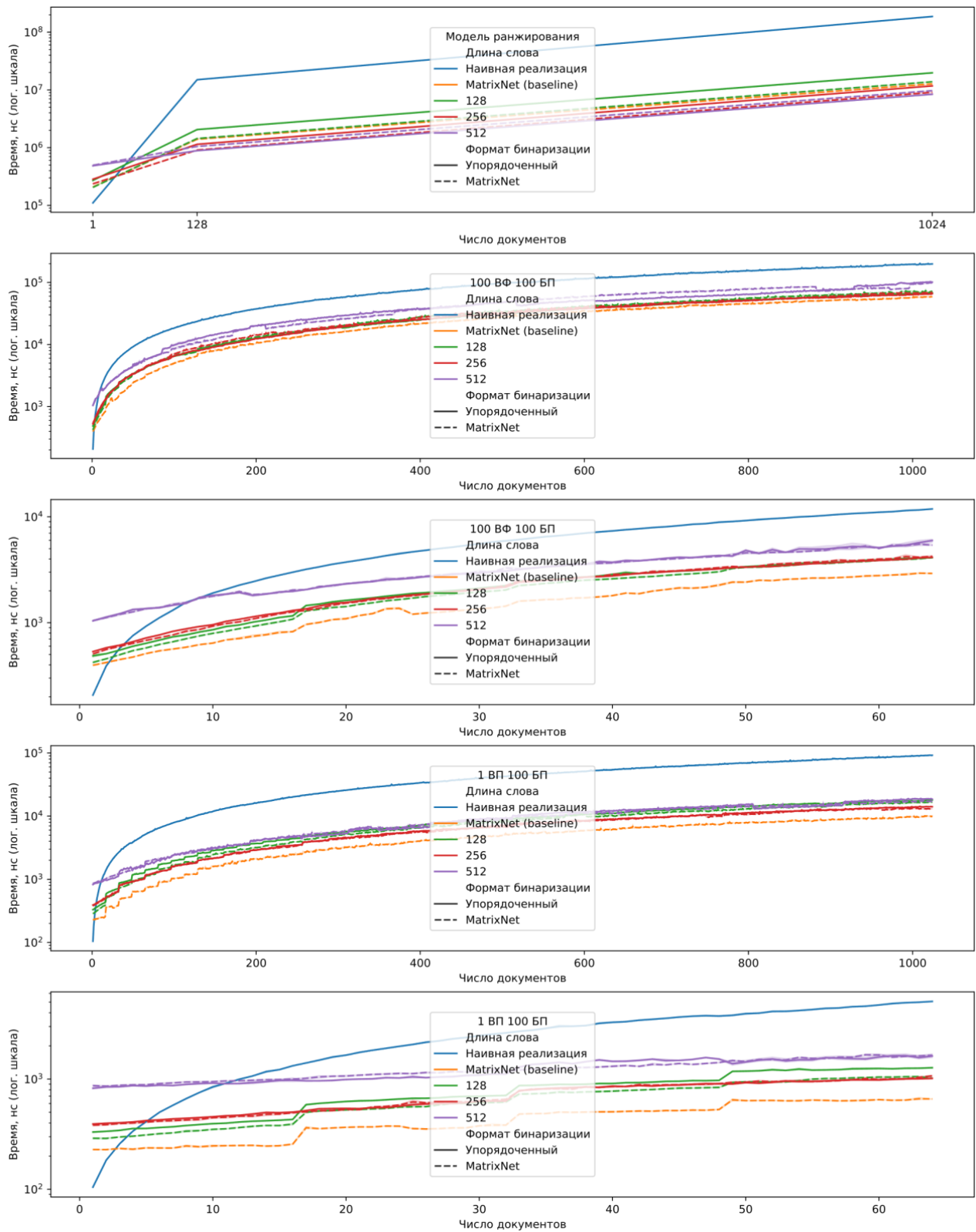

*Рисунок 5.1 Зависимость времени (логарифмическая шкала) от количества документов. Модель ранжирования – настоящая продуктовая модель, ВФ – вещественные факторы, БП – бинарные признаки. Посчитано для одного потока*

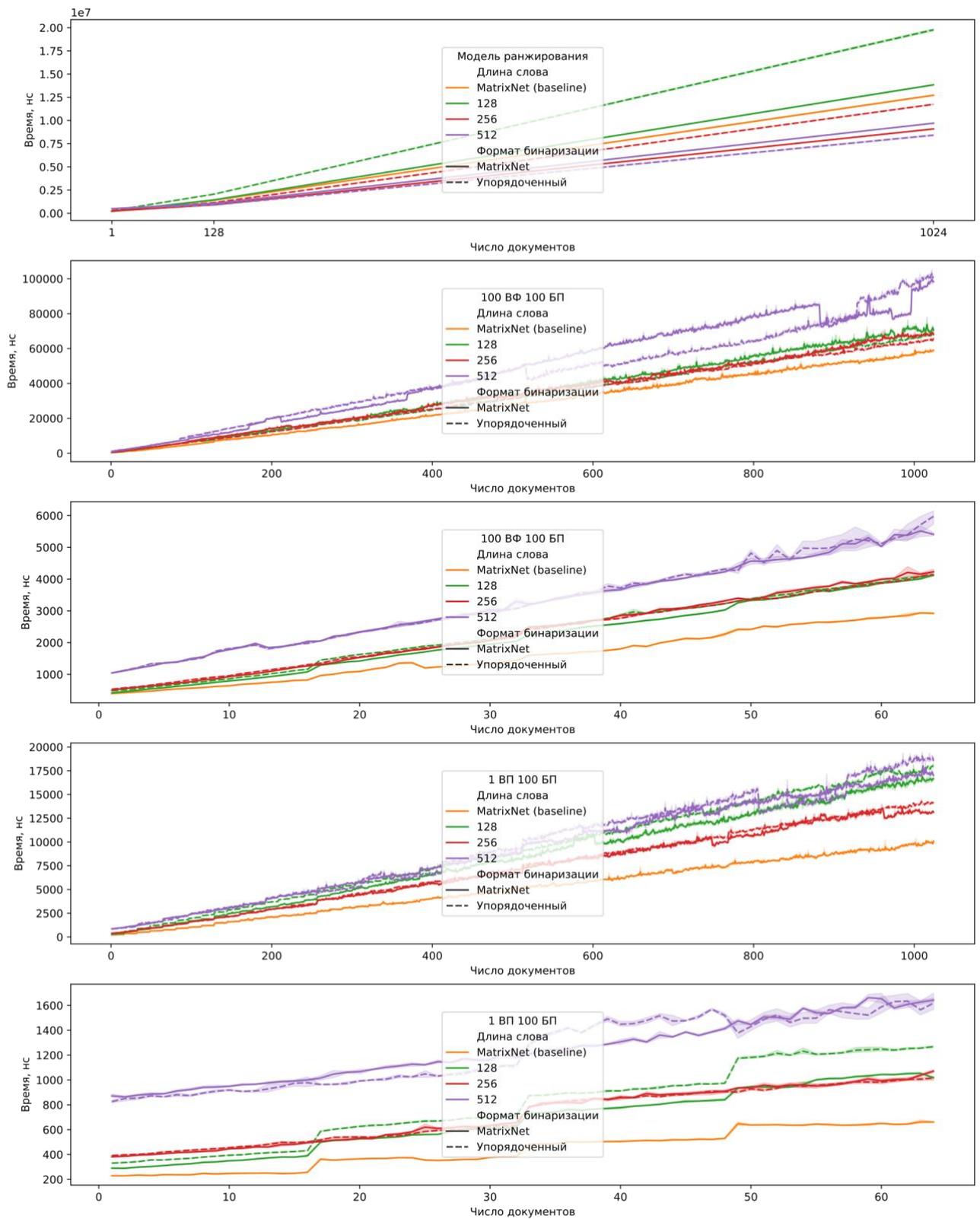

*Рисунок 5.2 Зависимость времени от количества документов. Наивная реализация исключена. Модель ранжирования – настоящая продуктовая модель, ВФ – вещественные факторы, БП – бинарные признаки. Посчитано для одного потока*

*Таблица 5.1 Замеры бинаризации 1024 документов. Модель ранжирования – настоящая продуктовая модель. Отклонение равно $(\text{new} - \text{old})/\text{old}$.*

| Алгоритм | Потоки | Модель ранжирования | | 100 факторов 100 б. признаков | | 1 фактор 100 б. признаков | |
| --- | --- | --- | --- | --- | --- | --- | --- |
| | | Время, нс | Отклонение от базовой | Время, нс | Отклонение от базовой | Время, нс | Отклонение от базовой |
| Наивная реализация | 1 | 186829000 | 13,582 | 199358 | 2,310 | 91551 | 7,952 |
| | 4 | 172059000 | | 248793 | | 162556 | |
| | 8 | 166903000 | | 293115 | | 207002 | |
| MatrixNet | 1 | 12812400 | 0,000 | 60233 | 0,000 | 10226 | 0,000 |
| | 4 | 13516100 | | 58925 | | 10592 | |
| | 8 | 14132600 | | 61139 | | 11260 | |
| AVX-128 | 1 | 13899100 | 0,085 | 71050 | 0,180 | 16876 | 0,650 |
| | 4 | 15117700 | | 73527 | | 18063 | |
| | 8 | 15910400 | | 76710 | | 18265 | |
| AVX-256 | 1 | 9058690 | -0,293 | 69560 | 0,155 | 13123 | 0,283 |
| | 4 | 10772500 | | 71373 | | 14035 | |
| | 8 | 12287400 | | 73086 | | 14994 | |
| AVX-512 | 1 | 9729130 | -0,241 | 98138 | 0,629 | 16854 | 0,648 |
| | 4 | 11550400 | | 97045 | | 19498 | |
| | 8 | 13127700 | | 101416 | | 20320 | |
| AVX-128 Упорядоченный | 1 | 19594300 | 0,529 | 67654 | 0,123 | 18344 | 0,794 |
| | 4 | 21163600 | | 67270 | | 18466 | |
| | 8 | 22064900 | | 68015 | | 19068 | |
| AVX-256 Упорядоченный | 1 | 11722800 | -0,085 | 65091 | 0,081 | 14171 | 0,386 |
| | 4 | 13346600 | | 66427 | | 14336 | |
| | 8 | 14618500 | | 66361 | | 14761 | |
| AVX-512 Упорядоченный | 1 | **8308580** | **-0,352** | 100234 | 0,664 | 18644 | 0,823 |
| | 4 | 10053100 | | 102864 | | 18461 | |
| | 8 | 11551400 | | 105786 | | 18960 | |
| AVX-128 Транспонирован. | 1 | 12119600 | -0,054 | 13121 | -0,782 | 7102 | -0,306 |
| | 4 | 12588300 | | 12810 | | 7047 | |
| | 8 | 13788900 | | 13593 | | 7288 | |
| AVX-256 Транспонирован. | 1 | 6643130 | -0,482 | 6776 | -0,888 | 3638 | -0,644 |
| | 4 | 7305860 | | 7169 | | 4108 | |
| | 8 | 8209680 | | 7498 | | 4417 | |
| AVX-512 Транспонирован. | 1 | 6997250 | -0,454 | 5084 | -0,916 | 4150 | -0,594 |
| | 4 | 7449790 | | 5313 | | 4266 | |
| | 8 | 8096370 | | 5620 | | 4585 | |

*Таблица 5.1 (продолжение)*

| | | | | | | | |
|---|---|---|---|---|---|---|---|
| AVX-128 | | | | | | | |
| Упорядоченный | 1 | 16881700 | 0,318 | 16065 | -0,733 | 10150 | -0,008 |
| Транспонирован. | 4 | 17627800 | | 16176 | | 10212 | |
| | 8 | 18952300 | | 16180 | | 10963 | |
| AVX-256 | | | | | | | |
| Упорядоченный | 1 | 8548370 | -0,333 | 8000 | -0,867 | 5094 | -0,502 |
| Транспонирован. | 4 | 9103120 | | 8330 | | 5310 | |
| | 8 | 9822450 | | 8584 | | 5338 | |
| AVX-512 | | | | | | | |
| Упорядоченный | 1 | **5041280** | **-0,607** | **4506** | **-0,925** | **2849** | **-0,721** |
| Транспонирован. | 4 | 5688940 | | 4814 | | 2938 | |
| | 8 | 6302770 | | 4948 | | 3139 | |

Видно, что для одного документа наивная реализация является наиболее эффективной, что ожидаемо, все улучшения в SIMD у нас достигаются за счёт обработки документов группами и подгруппами. Для большего числа документов уже лучше использовать SIMD. Эффект от подгрупп размера 16, 32 или 64 хорошо виден на графике модели со 100 бинарными признаками, но только одним вещественным фактором. На модели со 100 бинарными и 100 вещественными из-за расходов на загрузку вещественных факторов, этого эффекта не видно.

На транспонированной версии лучше виден эффект от SIMD инструкций: при наличии AVX-2, можно использовать AVX-256 формат, который даёт ускорение в 1.5 раза по сравнению с AVX-128. При наличии поддержки AVX-512F, можно использовать AVX-512 упорядоченный формат, который даёт ускорение в 2.5 раза по сравнению с AVX-128.

На не транспонированном же из-за долгой загрузки из памяти разница между алгоритмами менее заметна. Более того, на синтетических моделях MatrixNet реализация немного быстрее других, но ожидалось, что она будет на уровне AVX-128, так как там почти такой же алгоритм транспонирования.

Видно, что есть замедления от увеличения числа потоков, но на транспонированных и на наивной реализации, они менее заметны, что говорит о связи с промахами процессорного кэша.

## 6. Применение деревьев

### 6.1. Наивная реализация

На вход подаются бинарные признаки в одном из форматов, нужно выписать релевантности для всех документов.

Модель содержит три массива, относящихся к применению и задающих:

- число деревьев каждой из высот. Обозначение: $STC$. Деревья могут быть высоты от $0$ (ответ дерева константен) до $8$ (ответ дерева зависит от $8$ бинарных признаков).
- индексы бинарных признаков для каждого дерева. Массив состоит из целочисленных беззнаковых 32-битных чисел, длина равна $\sum_{i=0}^{9} i \cdot STC_i$.
- ответы в листьях для каждого дерева. Массив состоит из целочисленных беззнаковых 32-битных чисел, длина равна $\sum_{i=0}^{9} 2^i \cdot STC_i$.

Для каждого документа считаем сумму ответов деревьев в беззнаковом 64-битном числе. Рассматриваем деревья по неубыванию высоты. Поддерживаем сумму высот и сумму вторых степеней высот уже рассмотренных деревьев. Для очередного дерева высоты $n$ формируем $n$-битное число, где на смещении $i$ стоит бит, соответствующий бинарному признаку документа, индекс которого находится в массиве модели со смещением, равным сумме высот рассмотренных деревьев плюс $i$. Ответ для дерева берётся из массива ответов со смещением, равным сумме вторых степеней высот рассмотренных деревьев плюс полученное $n$-битное число.

Итоговый ответ для каждого документа получается по формуле $(R - 2^{31}) \cdot S + B$, где $B$ – полученная сумма, а $S$ и $B$ – вещественные параметры модели.

## 6.2. MatrixNet

В MatrixNet применение документов также выполняется над группами и подгруппами и разделено на две части: построение индексов и загрузка ответов.

При построении индексов для каждой группы, перебираются деревья так же, как в наивном алгоритме. Для каждого дерева по очереди рассматриваются бинарные признаки дерева. Для каждого бинарного признака загружаются в регистр 128 бит соответствующей ему бинаризации. Можно заметить, что в этой бинаризации $i$-ый бит в $j$-ом байте соответствует $j$-ому документу в $i$-ой подгруппе. Для каждого бинарного признака и для каждой подгруппы бинаризация сдвигается влево или вправо так, чтобы биты, соответствующие этой подгруппе, стояли на позициях, соответствующих позиции этого бинарного признака в текущем дереве, затем выполняются конъюнкция с маской, в который в каждом байте единичный бит стоит только на позиции этого бинарного признака в текущем дереве, чтобы обнулить биты других подгрупп. Наконец, для каждой подгруппы считается дизъюнкция всех таких конструкций.

Для каждой подгруппы получается 128-битное число, в каждом байте которого записан индекс ответа в дереве для одного из документов. Эти индексы сохраняются и после каждого четвёртого построения индексов, производится загрузка 32-битных ответов в SSE регистры как 64-битных и суммирование. Ответы для деревьев складываются для четырёх деревьев за раз. Вероятно, это для того, чтобы реже загружать промежуточные суммы.

Итоговый ответ считается так же, как в наивной реализации.

## 6.3. AVX реализация

Никаких проблем увеличение размера регистра не вызывает, все инструкции, используемые в MatrixNet реализации есть для всех размеров регистров. Единственное отличие связано с форматом бинаризации: Документы одной подгруппы лежат в бинаризации не по порядку, соответственно в конце их нужно поменять местами.

### 6.4. Упорядоченный формат

Здесь отличие заключается в том, что биты одной подгруппы лежат подряд, поэтому с одной стороны можно не загружать лишние биты, если подгрупп меньше 8, но с другой нужно превратить плотную запись в запись, где соседние значащие биты стоят на расстоянии 8. Для этого используется инструкция *vpbroadcastb*.

### 6.5. Загрузка ответов по одному дереву

Чтобы оценить, действительно ли есть смысл загружать ответы из четырёх деревьев за раз, была написана реализация, где загружается только для одного дерева за раз.

### 6.6. Загрузка ответов с помощью SIMD

Поскольку нужные нам числа лежат в разных частях массива, а в SSE были только инструкции последовательной загрузки, для загрузки использовались обычные *mov* инструкции. Однако в AVX-2 была добавлена инструкция *vpgatherdd* для регистров размера 128 и 256, а позже она была расширена и для 512 бит. Эта инструкция позволяет по 32-битным индексам загрузить 32-битные числа. Соответственно сначала нужно разделить регистр с 8-битными индексами на 4 регистра с 32-битными, потом выполнить загрузку ответов, затем разделить каждый регистр с 32-битными ответами на два регистра с 64-битными, и выполнить сложение. Увеличивать разрядность можно с помощью сдвигов и инструкций класса *unpack*.

### 6.7. Последовательная загрузка ответов

Наиболее быстрым является последовательное чтение из памяти, не только потому что это позволяет заранее загружать данные в кэш без использования инструкций типа *prefetch*, но и потому что это позволяет вычитывать в регистр кэш-линию целиком, эффективно расходуя ресурсы шины [9].

На используемой модели ранжирования максимальная высота деревьев равна 6, что значит, что при использовании 512-битных регистров, достаточно всего $2^6 \cdot \frac{32}{512} = 4$ регистров, чтобы уместить весь массив ответов дерева. Для деревьев высоты 8 (максимальная высота, поддерживаемая MatrixNet) нужно будет $2^8 \cdot \frac{32}{512} = 16$ регистров.

Как мы помним, индексы для массива ответов у нас восьмибитные. Если ответы лежат в 512-битных регистрах, то первые 4 бита индекса задают номер регистра, а последние 4 бита – номер в регистре.

Для перестановки ответов можно использовать инструкцию *vpermd*. Она принимает 512-битный регистр, содержащий 32-битные числа, ещё один 512-битный регистр, содержащий 4-битные индексы, и opmask регистр, содержащий маску для ответа, и возвращает регистр, состоящий из 32-битных чисел, где каждое число либо равно нулю, если соответствующий ему бит маски не установлен, либо выбрано из первого регистра с помощью соответствующего индекса из второго, если соответствующий бит в третьем установлен.

С помощью конъюнкции, сдвигов и операций распаковки мы разделяем регистр с индексами на два, содержащих первые или последние 4 бита каждого индекса. Далее для каждого регистра, содержащего часть массива ответа, выполняется перестановка, с использованием последних 4 бит индексов в качестве индексов перестановки, и маски, получаемой сравнением на равенство номера регистра с первыми 4 битами. Взяв дизъюнкцию получившихся перестановок, мы получаем ответы дерева для $\frac{512}{32} = 16$ документов.

Дальше результат получается так же, как и при SIMD загрузке.

## 6.8. Результаты

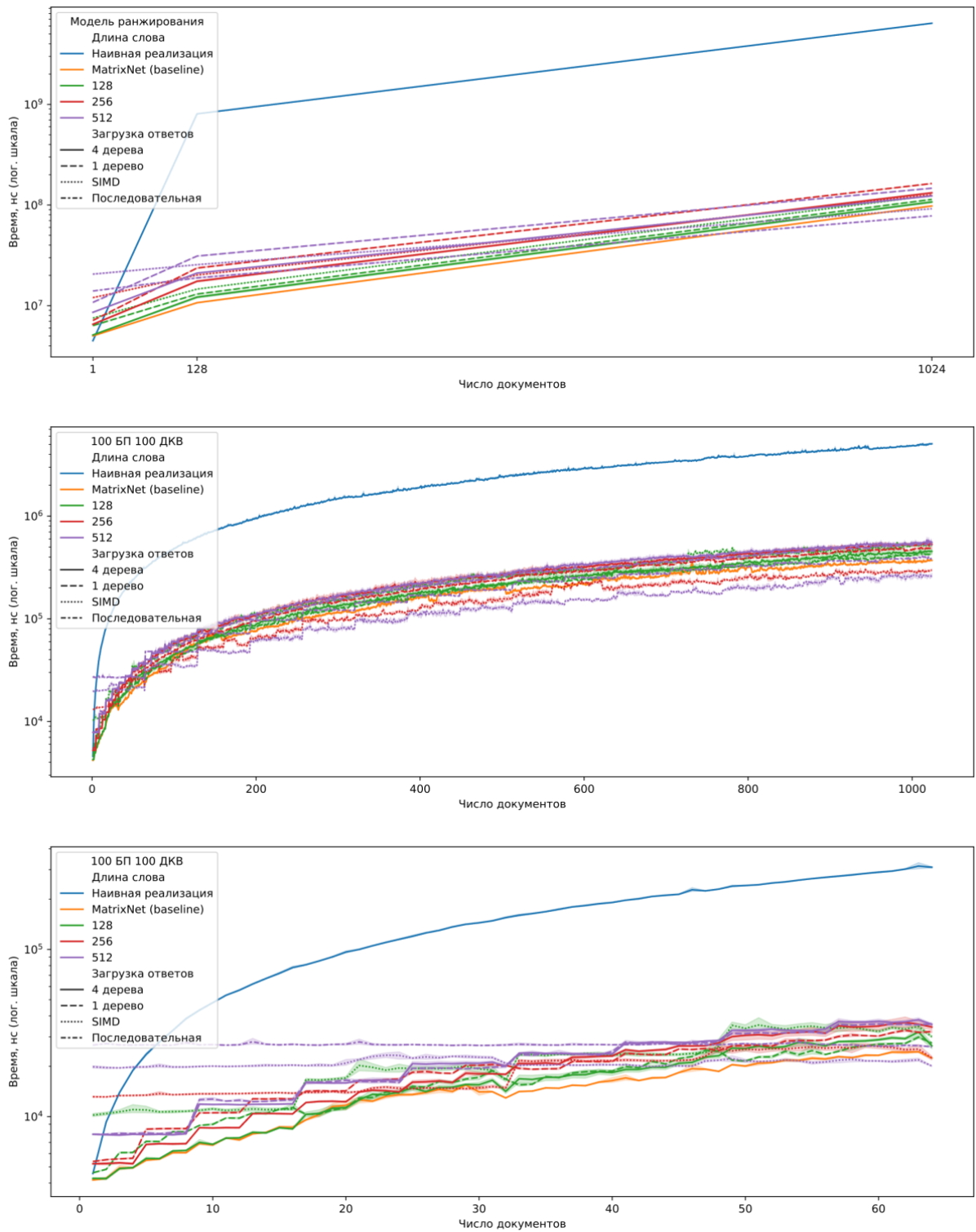

*Рисунок 6.1 Зависимость времени (логарифмическая шкала) от количества документов. Error bar виден, потому что Ordered и MatrixNet форматы объединены в одну линию. Модель ранжирования – настоящая продуктовая модель, БП – бинарные признаки, ДКВ – деревья каждой из высот от 0 до 9. Посчитано для одного потока*

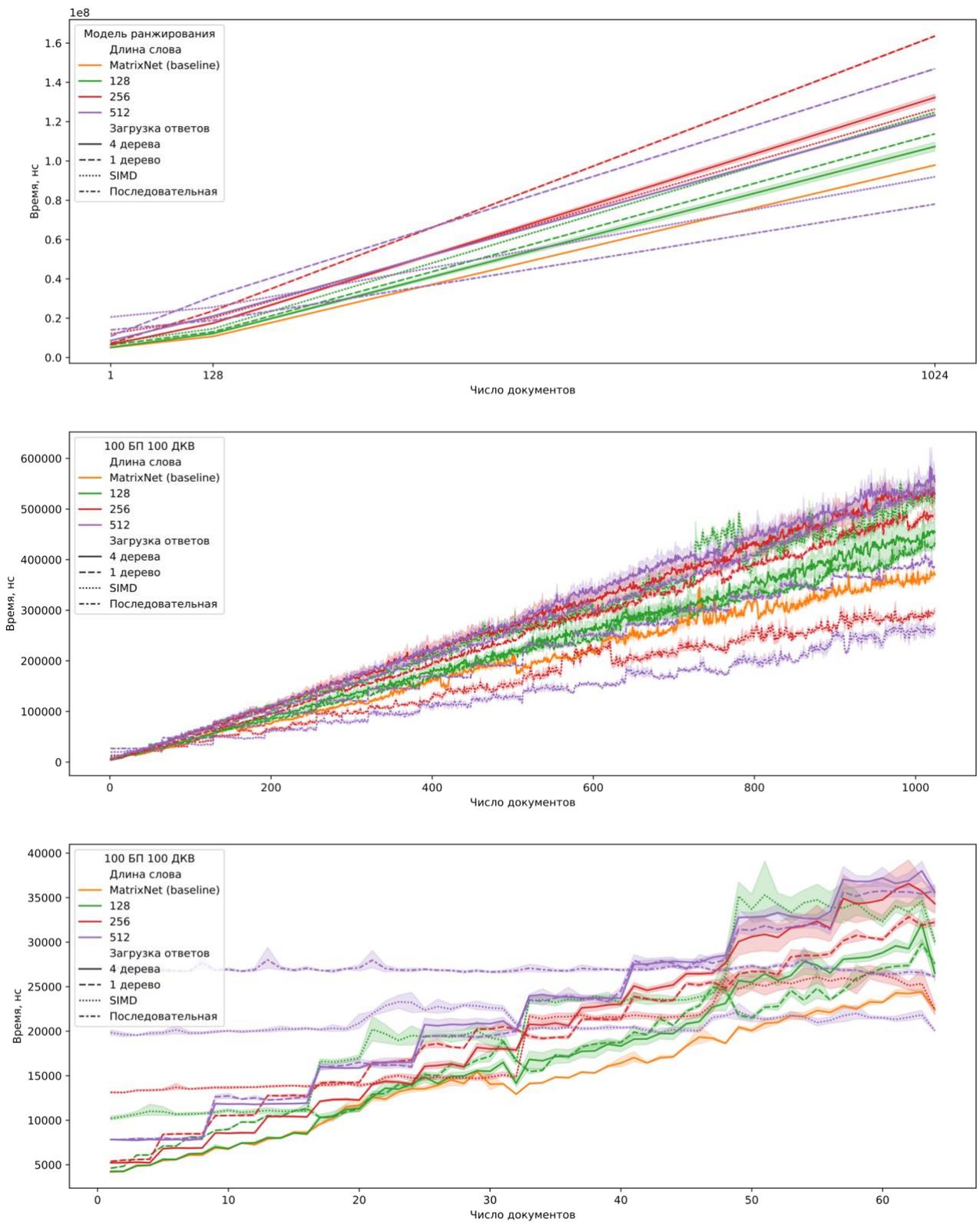

Рисунок 6.2 Зависимость времени от количества документов. Наивная реализация исключена. Error bar виден, потому что Ordered и MatrixNet форматы объединены в одну линию. Модель ранжирования – настоящая продуктовая модель, БП – бинарные признаки, ДКВ – деревья каждой из высот от 0 до 9. Посчитано для одного потока

*Таблица 6.1 Замеры бинаризации 1024 документов. Модель ранжирования – настоящая продуктовая модель. Отклонение равно (new − old)/old.*

| algo | Потоки | Модель ранжирования | | 100 б. признаков 100 деревьев каждой высоты | |
|---|---|---|---|---|---|
| | | Время, нс | Отклонение от базовой | Время, нс | Отклонение от базовой |
| Наивная реализация | 1 | 6355150000 | 64,051 | 5120790 | 12,654 |
| | 4 | 8449170000 | | 5415430 | |
| | 8 | 9275170000 | | 5549590 | |
| MatrixNet | 1 | 97694600 | 0,000 | 375039 | 0,000 |
| | 4 | 103370000 | | 430528 | |
| | 8 | 108141000 | | 471433 | |
| AVX-128 | 1 | 104663000 | 0,071 | 415817 | 0,109 |
| | 4 | 110418000 | | 481743 | |
| | 8 | 114596000 | | 509553 | |
| AVX-256 | 1 | 129995000 | 0,331 | 503402 | 0,342 |
| | 4 | 133262000 | | 529760 | |
| | 8 | 136133000 | | 549996 | |
| AVX-512 | 1 | 122296000 | 0,252 | 545096 | 0,453 |
| | 4 | 129917000 | | 555752 | |
| | 8 | 137790000 | | 635266 | |
| AVX-128 Упорядоченный | 1 | 111352000 | 0,140 | 484664 | 0,292 |
| | 4 | 114953000 | | 557670 | |
| | 8 | 119013000 | | 583892 | |
| AVX-256 Упорядоченный | 1 | 135596000 | 0,388 | 558249 | 0,489 |
| | 4 | 137645000 | | 693763 | |
| | 8 | 140493000 | | 709932 | |
| AVX-512 Упорядоченный | 1 | 124653000 | 0,276 | 573994 | 0,530 |
| | 4 | 133224000 | | 673478 | |
| | 8 | 140703000 | | 728730 | |
| AVX-128 Загрузка по одному дереву | 1 | 113717000 | 0,164 | 415711 | 0,108 |
| | 4 | 119744000 | | 475314 | |
| | 8 | 123683000 | | 520360 | |
| AVX-256 Загрузка по одному дереву | 1 | 162716000 | 0,666 | 475598 | 0,268 |
| | 4 | 167204000 | | 512149 | |
| | 8 | 170937000 | | 537123 | |
| AVX-512 Загрузка по одному дереву | 1 | 146532000 | 0,500 | 545526 | 0,455 |
| | 4 | 153348000 | | 570713 | |
| | 8 | 160500000 | | 587436 | |
| AVX-128 SIMD загрузка | 1 | 125132000 | 0,281 | 533925 | 0,424 |
| | 4 | 130598000 | | 574781 | |
| | 8 | 135023000 | | 625636 | |
| AVX-256 SIMD загрузка | 1 | 126591000 | 0,296 | 298041 | -0,205 |
| | 4 | 130086000 | | 346306 | |
| | 8 | 132232000 | | 375780 | |
| AVX-512 SIMD загрузка | 1 | 92326300 | -0,055 | **286979** | **-0,235** |
| | 4 | 96988400 | | 286908 | |
| | 8 | 101694000 | | 304738 | |
| AVX-512 Последовательная загрузка | 1 | **77788900** | -0,204 | 386945 | 0,031 |
| | 4 | 82435500 | | 415371 | |
| | 8 | 88007200 | | 437215 | |

Видно, что увеличение размера регистра без изменения алгоритма не даёт преимущества, в применении нет вычислительно сложных операций, но много операций загрузки из памяти. Разницы, в каком из форматов поступают бинаризованные данные при большом числе документов так же нет. Утилита perf показывает, что большую часть времени составляет загрузка из памяти.

Зато есть разница, как загружать ответы деревьев из памяти. Нет большой разницы, грузить по одному или по 4 дерева за раз, но загрузка с помощью SIMD инструкций, как и последовательная загрузка, ускоряют. При SIMD загрузке на синтетической модели со 100 деревьями каждой высот видно ускорение на 24% с AVX-512 по сравнению с MatrixNet, но при этом на настоящей модели ранжирования время сокращается только на 6%. Так же видно, что при SIMD загрузке имеет значение размер регистра, и при небольшом количестве документов его использование не эффективно.

При последовательной загрузке на поисковой модели видно ускорение на 20%, но для синтетической модели ускорения нет, что вероятно связано с тем, что в синтетической модели деревья глубже и нужно слишком много операций для перестановки ответов. Как и SIMD загрузку, последовательную загрузку имеет смысл использовать только при большом числе документов.

Замедление при увеличении числа потоков небольшое и так же объясняется кэшем процессора.

## 7. Заключение

В работе представлены несколько алгоритмов бинаризации и применения. Из их числа можно взять любой алгоритм бинаризации и любой алгоритм применения и использовать их вместе. Возможно, только в конце придётся поменять порядок документов. Делать выбор можно опираясь на поддерживаемые процессором расширения и замеры скорости на конкретной модели и конкретном числе документов.

В работе было показано, что скорость работы с памятью и кэшем часто влияют на результат больше, чем скорость самих вычислений. Соответственно, для бинаризации, возможно, лучше не делать транспонирование, но тогда больше раз придётся вычитывать пороги, а также, не понятно, как решать проблему с тем, что в одном регистре будут лежать разные вещественные факторы, для которых нужно выполнить разное число сравнений с разными порогами, и непонятно, как сделать подходящий формат бинаризации, по которому потом можно будет быстро применять.

Для применения документов имеет смысл поэкспериментировать с порядком бинарных признаков и деревьев, так как от этого может зависеть число промахов кэша и скорость работы.

## Библиографический список